%% file: main-4637-He.tex
\documentclass[11pt,a4paper]{article}
\usepackage{times,latexsym}
\usepackage{microtype}
\usepackage{url}
\usepackage[T1]{fontenc}

\usepackage[acceptedWithA]{tacl2021v1}

\usepackage{xspace,mfirstuc,tabulary}

\newif\iftaclinstructions
\taclinstructionsfalse 
\iftaclinstructions
\renewcommand{\confidential}{}
\renewcommand{\anonsubtext}{(No author info supplied here, for consistency with
TACL-submission anonymization requirements)}
\newcommand{\instr}
\fi

\iftaclpubformat 

\else

\fi

\usepackage{multirow}
\usepackage{caption}
\usepackage{subcaption}
\usepackage{amsmath}
\usepackage{booktabs}       
\usepackage{graphicx}
\usepackage[ruled,vlined]{algorithm2e}
\usepackage[hang,flushmargin]{footmisc}  
\usepackage{soul}

\newcommand{\LN}{\linebreak\noindent}    
\newcommand{\POS}{\texttt{POS}}
\newcommand{\NER}{\texttt{NER}}
\newcommand{\DEP}{\texttt{DEP}}
\newcommand{\CON}{\texttt{CON}}

\title{Unleashing the True Potential of Sequence-to-Sequence Models\\ for Sequence Tagging and Structure Parsing}

\author{Han He \\
  Department of Computer Science \\
  Emory University \\
  Atlanta, GA 30322 USA \\
  \texttt{han.he@emory.edu} \\\And
  Jinho D. Choi \\
  Department of Computer Science \\
  Emory University \\
  Atlanta, GA 30322 USA \\
  \texttt{jinho.choi@emory.edu} \\}

\date{}

\begin{document}
\maketitle

\input{tex/abstract}
\input{tex/introduction}
\input{tex/related-work}
\input{tex/schema}
\input{tex/decoding-strategies}

\input{tex/experiments}
\input{tex/analysis}
\input{tex/conclusion}

\section*{Acknowledgments}

We would like to thank Emily Pitler, Cindy Robinson, Ani Nenkova, and the anonymous TACL reviewers for their insightful and thoughtful feedback on the early drafts of this paper.

\bibliography{references}
\bibliographystyle{acl_natbib}

\end{document}

%% file: tex/abstract.tex
\begin{abstract}

Sequence-to-Sequence (S2S) models have achieved remarkable success on various text generation tasks.
However, learning complex structures with S2S models remains challenging as external neural modules and additional lexicons are often supplemented to predict non-textual outputs.
We present a systematic study of S2S modeling using contained decoding on four core tasks: part-of-speech tagging, named entity recognition, constituency and dependency parsing, to develop efficient exploitation methods costing zero extra parameters.
In particular, 3 lexically diverse linearization schemas and corresponding constrained decoding methods are designed and evaluated.
Experiments show that although more lexicalized schemas yield longer output sequences that require heavier training, their sequences being closer to natural language makes them easier to learn.
Moreover, S2S models using our constrained decoding outperform other S2S approaches using external resources.
Our best models perform better than or comparably to the state-of-the-art for all 4 tasks, lighting a promise for S2S models to generate non-sequential structures.


\end{abstract}

%% file: tex/introduction.tex
\section{Introduction}
\label{sec:introduction}

Sequence-to-Sequence (S2S) models pretrained for language modeling (PLM) and denoising objectives have been successful on a wide range of NLP tasks where both inputs and outputs are sequences \cite{radford2019language,JMLR:v21:20-074, lewis-etal-2020-bart, NEURIPS2020_1457c0d6}. 
However, for non-sequential outputs like trees and graphs, a procedure called linearization is often required to flatten them into ordinary sequences \cite{li-etal-2018-seq2seq, fernandez-gonzalez-gomez-rodriguez-2020-enriched,yan-etal-2021-unified-generative, bevilacqua2021one, he-choi-2021-levi}, where labels in non-sequential structures are mapped heuristically as individual tokens in sequences, and numerical properties like indices are either predicted using an external decoder such as Pointer Networks \cite{vinyals2015pointer} or cast to additional tokens in the vocabulary. 
While these methods are found to be effective, we hypothesize that S2S models can learn complex structures without adapting such patches.

To challenge the limit of S2S modeling,  BART \cite{lewis-etal-2020-bart} is finetuned on four tasks without extra decoders: part-of-speech tagging (\POS), named entity recognition (\NER), constituency parsing (\CON), and dependency parsing (\DEP).
Three novel linearization schemas are introduced for each task: label sequence (\texttt{LS}), label with text (\texttt{LT}), and prompt (\texttt{PT}).
\texttt{LS} to \texttt{PT} feature an increasing number of lexicons and a decreasing number of labels,\LN which are not in the vocabulary (Section~\ref{sec:schemas}). 
Every schema is equipped with a constrained decoding algorithm searching over valid sequences (Section~\ref{sec:decoding-strategies}).

Our experiments on three popular datasets depict that S2S models can learn these linguistic structures without external resources such as index tokens or Pointer Networks.
Our best models perform on par with or better than the other state-of-the-art models for all four tasks (Section~\ref{sec:experiments}).
Finally, a detailed analysis is provided to compare the distinctive natures of our proposed schemas (Section~\ref{sec:analysis}).\footnote{All our resources including source codes are publicly available: \url{https://github.com/emorynlp/seq2seq-corenlp}}

%% file: tex/related-work.tex
\vspace{-0.3em}
\section{Related Work}
\label{sec:related-work}

S2S \cite{sutskever2014sequence} architectures have been effective on many sequential modeling tasks.
Conventionally, S2S is implemented as an encoder and decoder pair, where the encoder learns input representations used to generate the output sequence via the decoder. 
Since the input sequence can be very long, attention mechanisms \cite{DBLP:journals/corr/BahdanauCB14, vaswani2017attention} focusing on particular positions are often augmented to the basic architecture. 
With transfer-learning, S2S models pretrained on large unlabeled corpora have risen to a diversity of new approaches that convert language problems into a text-to-text format \cite{akbik:18a, lewis-etal-2020-bart, radford2019language,JMLR:v21:20-074, NEURIPS2020_1457c0d6}. 
Among them, tasks most related to our work are linguistic structure predictions using S2S, \POS, \NER, \DEP, and \CON.

\POS{} has been commonly tackled as a sequence tagging task, where the input and output sequences have equal lengths. 
S2S, on the other hand, does not enjoy such constraints as the output sequence can be arbitrarily long. Therefore, S2S is not as popular as sequence tagging for \POS. Prevailing neural architectures for \POS{} are often built on top of a neural sequence tagger with rich embeddings \cite{bohnet:18a, akbik:18a} and Conditional Random Fields \cite{LaffertyMP01}.

\NER{} has been cast to a neural sequence tagging task using the IOB notation \cite{lample-etal-2016-neural} over the years, which benefits most from contextual word embeddings \cite{devlin-etal-2019-bert, wang-etal-2021-automated}. 
Early S2S-based works cast NER to a text-to-IOB transduction problem \cite{chen-moschitti-2018-learning, strakova-etal-2019-neural, zhu2020fine}, which is included as a baseline schema in Section~\ref{sec:ner-schema}. 
\citet{yan-etal-2021-unified-generative} augment Pointer Networks to generate numerical entity spans, which we refrain to use because the focus of this work is purely on the S2S itself. Most recently, \citet{cui-etal-2021-template} propose the first template prompting to query all possible spans against a S2S language model, which is highly simplified into a one-pass generation in our \texttt{PT} schema. Instead of directly prompting for the entity type, \citet{chen-etal-2022-shot} propose to generate its concepts first then its type later. Their two-step generation is tailored for few-shot learning, orthogonal to our approach. Moreover, our prompt approach does not rely on non-textual tokens as they do.

\CON{} is a more established task for S2S models since the bracketed constituency tree is naturally a linearized sequence. Top-down tree linearizations based on brackets \cite{NIPS2015_277281aa} or shift-reduce actions \cite{sagae-lavie-2005-classifier} rely on a strong encoder over the sentence while bottom-up ones \cite{zhu-etal-2013-fast,ma2017deterministic} can utilize rich features from readily built partial parses. Recently, the in-order traversal has proved superior to bottom-up and top-down in both transition \cite{liu-zhang-2017-order} and S2S \cite{fernandez-gonzalez-gomez-rodriguez-2020-enriched} constituency parsing. 
Most recently, a Pointer Networks augmented approach \cite{yang-tu-2022-bottom} is ranked top among S2S approaches.
Since we are interested in the potential of S2S models without patches, a naive bottom-up baseline and its novel upgrades are studied in Section~\ref{sec:con-schema}. 

\DEP{} has been underexplored as S2S due to the linearization complexity. 
The first S2S work maps a sentence to a sequence of source sentence words interleaved with the arc-standard, reduce-actions in its parse \cite{wiseman-rush-2016-sequence}, which is adopted as our \texttt{LT} baseline in Section~\ref{sec:dep-schema}. \citet{zhang-etal-2017-stack} introduce a stack-based multi-layer attention mechanism to leverage structural linguistics information from the decoding stack in arc-standard parsing. Arc-standard is also used in our \texttt{LS} baseline, however, we use no such extra layers. 
Apart from transition parsing, \citet{li-etal-2018-seq2seq} directly predict the relative head position instead of the transition. This schema is later extended to multilingual and multitasking by \citet{choudhary-oriordan-2021-end}.
Their encoder and decoder use different vocabularies, while in our \texttt{PT} setting, we re-use the vocabulary in the S2S language model.

S2S appears to be more prevailing for semantic parsing due to two reasons. First, synchronous context-free grammar bridges the gap between natural text and meaning representation for S2S. It has been employed to obtain silver annotations \cite{jia-liang-2016-data}, and to generate canonical natural language paraphrases that are easier to learn for S2S \cite{shin-etal-2021-constrained}. This trend of insights viewing semantic parsing as prompt-guided generation \cite{hashimoto2018retrieve} and paraphrasing \cite{berant-liang-2014-semantic} has also inspired our design of \texttt{PT}. Second, the flexible input/output format of S2S facilitates joint learning of semantic parsing and generation.  Latent variable sharing \cite{tseng-etal-2020-generative} and unified pretraining \cite{bai-etal-2022-graph} are two representative joint modeling approaches, which could be augmented with our idea of \texttt{PT} schema as a potentially more effective linearization.

Our finding that core NLP tasks can be solved using \texttt{LT} overlaps with the Translation between Augmented Natural Languages \cite{paolini2021structured}. However, we take one step further to study the impacts of textual tokens in schema design choices. Our constrained decoding is similar to existing works \cite{hokamp-liu-2017-lexically, deutsch-etal-2019-general,shin-etal-2021-constrained}. We craft constrained decoding algorithms for our proposed schemas and provide a systematic ablation study in Section~\ref{sec:abulation}.

%% file: tex/schema.tex
\begin{table*}[ht!]
    \centering\small{ \resizebox{\textwidth}{!}{
    \begin{tabular}{l|l} 
    \toprule
    \multicolumn{2}{c}{\bf Part-of-Speech Tagging} \\
    \midrule
    \multicolumn{2}{c}{\includegraphics[width=0.6\textwidth]{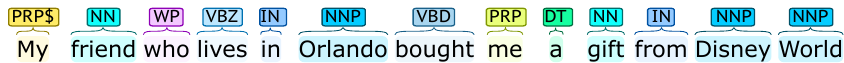}} \\
    \midrule
    \texttt{LS} & \texttt{PRP\$} \texttt{NN} \texttt{WP} \texttt{VBZ} \texttt{IN} \texttt{NNP} \texttt{VBD} \texttt{PRP} \texttt{DT} \texttt{NN} \texttt{IN} \texttt{NNP} \texttt{NNP} \\
    \midrule
    \multirow{2}{*}{\texttt{LT}}
     & My\texttt{/PRP\$} friend\texttt{/NN} who\texttt{/WP} lives\texttt{/VBZ} in\texttt{/IN} Orlando\texttt{/NNP} \\
     & bought\texttt{/VBD} me\texttt{/PRP} a\texttt{/DT} gift\texttt{/NN} from\texttt{/IN} Disney\texttt{/NNP} World\texttt{/NNP} \\
    \midrule
    \multirow{3}{*}{\texttt{PT}}
     & ``My'' is a \textit{possessive pronoun}; ``friend'' is a \textit{noun}; ``who'' is a \textit{wh-pronoun}; ``lives'' is a \textit{3rd-person present verb}; \\
     & ``in'' is a \textit{preposition}; ``Orlando'' is a \textit{proper noun}; ``bought'' is a \textit{past verb}; ``me'' is a \textit{personal pronoun};\\
     & ``a'' is a \textit{determiner}; ``gift'' is a \textit{noun}; ``from'' is a \textit{preposition}; ``Disney'' is a \textit{proper noun}; ``World'' is a \textit{proper noun}. \\
    \toprule
    \multicolumn{2}{c}{\bf Named Entity Recognition} \\
    \midrule
    \multicolumn{2}{c}{\includegraphics[width=0.55\textwidth]{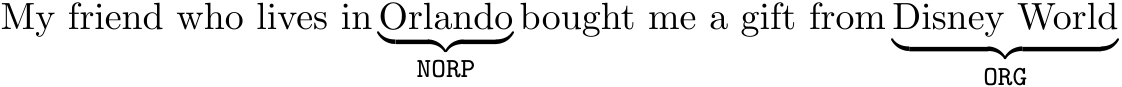}} \\
    \midrule
    %
    \texttt{LS}     & \texttt{O} \texttt{O} \texttt{O} \texttt{O} \texttt{O} \texttt{S-NORP} \texttt{O} \texttt{O} \texttt{O} \texttt{O} \texttt{O} \texttt{B-ORG} \texttt{E-ORG} \\
    \midrule
    \texttt{LT} & My friend who lives in \texttt{<NORP>}Orlando\texttt{</NORP>} bought me a gift from \texttt{<ORG>}Disney World\texttt{</ORG>} \\
    \midrule
    \texttt{PT}     & ``Orlando'' is a \textit{geopolitical entity}; ``Disney World'' is an \textit{organization}. \\
    \toprule
    \multicolumn{2}{c}{\bf Constituency Parsing} \\
    \midrule
    \multicolumn{2}{c}{\includegraphics[width=0.7\textwidth]{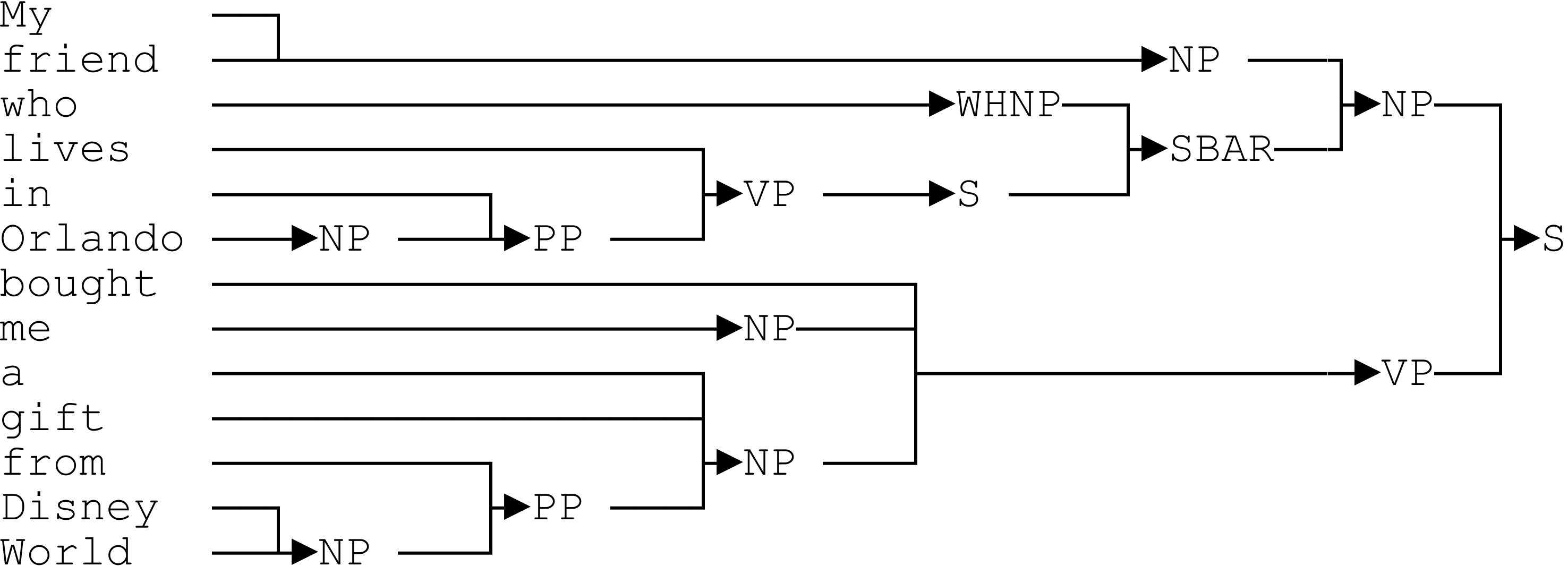}} \\
    \midrule
    \multirow{2}{*}{\texttt{LS}}
     & \texttt{N-S} \texttt{N-NP} \texttt{N-NP} \texttt{SH} \texttt{SH} \texttt{RE} \texttt{N-SBAR} \texttt{N-WHNP} \texttt{SH} \texttt{RE} \texttt{N-S} \texttt{N-VP} \texttt{SH} \texttt{N-PP} \texttt{SH} \texttt{N-NP} \texttt{SH} \texttt{RE} \texttt{RE} \texttt{RE} \texttt{RE} \texttt{RE} \texttt{RE} \\
     & \texttt{N-VP} \texttt{SH} \texttt{N-NP} \texttt{SH} \texttt{RE} \texttt{N-NP} \texttt{SH} \texttt{SH} \texttt{N-PP} \texttt{SH} \texttt{N-NP} \texttt{SH} \texttt{SH} \texttt{RE} \texttt{RE} \texttt{RE} \texttt{RE} \texttt{RE} \\
    \midrule
    \multirow{2}{*}{\texttt{LT}}
     & \texttt{(S} \texttt{(NP} \texttt{(NP} My friend\texttt{)} \texttt{(SBAR} \texttt{(WHNP} who\texttt{)} \texttt{(S} \texttt{(VP} lives \texttt{(PP} in \texttt{(NP} Orlando\texttt{))))))} \\
     & \texttt{(VP} bought \texttt{(NP} me\texttt{)} \texttt{(NP} a gift \texttt{(PP} from \texttt{(NP} Disney World\texttt{)))))} \\
    \midrule
    \multirow{5}{*}{\texttt{PT}}
     & The \textit{sentence} has a \textit{noun phrase} and a \textit{verb phrase}; The \textit{noun phrase} has the \textit{noun phrase} ``My friend'' and \\
     & the \textit{subordinating clause}, which has the \textit{wh-noun phrase} ``who'' and the \textit{clause}, which has the \textit{verb phrase}, \\
     & which has ``lives'' and the \textit{preposition phrase}, which has ``in'' and the \textit{noun phrase} ``Orlando''; \\
     & the \textit{verb phrase} has ``bought'' and the \textit{noun phrase} ``me'' and the \textit{noun phrase} ``a gift'' and the \textit{preposition phrase},\\
     & which has ``from'' and the \textit{noun phrase} ``Disney World''. \\
    \toprule
    \multicolumn{2}{c}{\bf Dependency Parsing} \\
    \midrule
    \multicolumn{2}{c}{\includegraphics[width=0.85\textwidth]{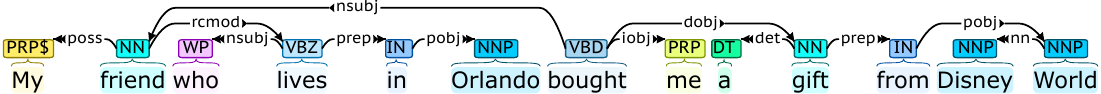}} \\
    \midrule
    \multirow{2}{*}{\texttt{LS}}
     & \texttt{SH} \texttt{SH} \texttt{<poss} \texttt{SH} \texttt{SH} \texttt{<nsubj} \texttt{SH} \texttt{SH} \texttt{<case} \texttt{>obl} \texttt{>relcl} \\
     &  \texttt{SH} \texttt{<nsubj} \texttt{SH} \texttt{>iobj} \texttt{SH} \texttt{SH} \texttt{<det} \texttt{SH} \texttt{SH} \texttt{SH} \texttt{<compound} \texttt{<case} \texttt{>nmod} \texttt{>obj} \\
    \midrule
    \multirow{2}{*}{\texttt{LT}}
     & My friend \texttt{<poss} who lives \texttt{<nsubj} in Orlando \texttt{<case} \texttt{>obl} \texttt{>relcl} \\
     & bought \texttt{<nsubj} me \texttt{>iobj} a gift \texttt{<det} from Disney World \texttt{<compound} \texttt{<case} \texttt{>nmod} \texttt{>obj} \\
    \midrule
    \multirow{4}{*}{\texttt{PT}}
     & ``My'' is a \textit{possessive modifier} of ``friend''; ``who'' is a \textit{nominal subject} of ``lives''; ``in'' is a \textit{case marker} of ``Orlando'';\\
     & ``lives'' has an \textit{oblique} ``Orlando''; ``friend'' has a \textit{relative clause} ``lives''; ``friend'' is a \textit{nominal subject} of ``bought'';\\
     & ``bought'' has an \textit{indirect object} ``me''; ``a'' is a \textit{determiner} of ``gift''; ``Disney'' is a \textit{compound word} of ``World'';\\
     & ``from'' is a \textit{case marker} of ``World''; ``gift'' has a \textit{nominal modifier} ``World''; ``bought'' has an \textit{object} ``gift''. \\
    \bottomrule
    \end{tabular}}
    }
    \caption{Schemas for the sentence ``My friend who lives in Orlando bought me a gift from Disney World''.  }
    \label{tab:schemas}
    \end{table*}

\section{Schemas}
\label{sec:schemas}

This section presents our output schemas for \POS, \NER, \CON{}, and \DEP{} in Table~\ref{tab:schemas}. 
For each task, 3 lexically diverse schemas are designed as follows to explore the best practice for structure learning.
First, \texttt{L}abel \texttt{S}equence (\texttt{LS}) is defined as a sequence of labels consisting of a finite set of task-related labels, that are merged into the S2S vocabulary, with zero text.
Second, \texttt{L}abel with \texttt{T}ext (\texttt{LT}) includes tokens from the input text on top of the labels such that it has a medium number of labels and text.
Third, \texttt{P}romp\texttt{T} (\texttt{PT}) gives a list of sentences describing the linguistic structure in natural language with no label. 
We hypothesize that the closer the output is to natural language, the more advantage the S2S takes from the PLM.



\subsection{Part-of-Speech Tagging (\POS)}

\paragraph{\texttt{LS}} \texttt{LS} defines the output as a sequence of \POS{} tags. 
Formally, given an input sentence of $n$ tokens $\mathbf{x}=\{x_1, x_2, \cdots, x_n\}$, its output is a tag sequence of the same length $\mathbf{y}^{\texttt{LS}}=\{y_1,y_2, \cdots,y_n\}$. 
Distinguished from sequence tagging, any \texttt{LS} output sequence is terminated by the ``end-of-sequence'' (\texttt{EOS}) token, which is omitted from $\mathbf{y}^{\texttt{LS}}$ for simplicity. 
Predicting \POS{} tags often depends on their neighbor contexts. 
We challenge that the autoregressive decoder of a S2S model can capture this dependency through self-attention.

\paragraph{\texttt{LT}} For \texttt{LT}, the token from the input is inserted before its corresponding tag. 
Formally, the output is defined $\mathbf{y}^{\texttt{LT}}=\{(x_1, y_1),(x_2, y_2), ..,(x_n,y_n)\}$. 
Both $\mathbf{x}$ and $\mathbf{y}$ are part of the output and the S2S model is trained to generate each pair sequentially.

\paragraph{\texttt{PT}} \texttt{PT} is a human-readable text describing the \POS{} sequence. Specifically, we use a phrase $\mathbf{y}^{\texttt{PT}}_i=$``$x_i$ is $y_i^{\prime}$'' for the $i$-th token, where $y_i^{\prime}$ is the definition of a \POS{} tag $y_i$, e.g., \texttt{a noun}. 
The final prompt is then the semicolon concatenation of all phrases: $\mathbf{y}^{\texttt{PT}}=\mathbf{y}^{\texttt{PT}}_1; \mathbf{y}^{\texttt{PT}}_2; \cdots; \mathbf{y}^{\texttt{PT}}_n$.

\subsection{Named Entity Recognition (\NER)}
\label{sec:ner-schema}

\paragraph{\texttt{LS}} \texttt{LT} of an input sentence comprising $n$ tokens $\mathbf{x}=\{x_1, x_2, \cdots, x_n\}$ is defined as the BIEOS tag sequence $\mathbf{y}^{\texttt{LS}}=\{y_1,y_2, \cdots,y_n\}$, which labels each token as the \texttt{B}eginning, \texttt{I}nside, \texttt{E}nd, \texttt{O}utside, or \texttt{S}ingle-token entity. 

\paragraph{\texttt{LT}} \texttt{LT} uses a pair of entity type labels to wrap each entity: $\mathbf{y}^{\texttt{LT}}=..  \texttt{B-}y_j, x_i, .., x_{i+k}, \texttt{E-}y_j, ..$, where $y_j$ is the type label of the $j$-th entity consisting of $k$ tokens. 

\paragraph{\texttt{PT}} \texttt{PT} is defined as a list of sentences describing each entity: $\mathbf{y}^{\texttt{PT}}_i=$``$x_i$ is $y_i^{\prime}$'', where $y_i^{\prime}$ is the definition of a \NER{} tag $y_i$, e.g., \texttt{a person}. Different from the prior prompt work \cite{cui-etal-2021-template}, our model generates all entities in one pass which is more efficient than their brute-force approach.

\subsection{Constituency Parsing (\CON)}
\label{sec:con-schema}

Schemas for \CON{} are developed on constituency trees pre-processed by removing the first level of non-terminals (\POS{} tags) and rewiring their children (tokens) to parents, e.g., \texttt{(NP (PRON My) (NOUN friend))} $\rightarrow$ \texttt{(NP My friend)}.

\paragraph{\texttt{LS}} \texttt{LS} is based on a top-down shift-reduce system consisting of a stack, a buffer, and a depth record $d$. Initially, the stack contains only the root constituent with label \texttt{TOP} and depth $0$; the buffer contains all tokens from the input sentence; $d$ is set to $1$. A \texttt{Node-X} (\texttt{N-X}) transition creates a new depth-$d$ non-terminal labeled with \texttt{X}, pushes it to the stack, and sets $d\leftarrow d+1$. A \texttt{Shift} (\texttt{SH}) transition removes the first token from the buffer and pushes it to the stack as a new terminal with depth $d$. A \texttt{Reduce} (\texttt{RE}) pops all elements with the same depth $d$ from the stack then make them the children of the top constituent of the stack, and it sets $d\leftarrow d-1$. The linearization of a constituency tree using our \texttt{LS} schema can be obtained by applying 3 string substitutions: replace each left bracket and the label \texttt{X} following it with a \texttt{Node-X}, replace terminals with \texttt{SH}, replace right brackets with \texttt{RE}.

\paragraph{\texttt{LT}} \texttt{LT} is derived by reverting all \texttt{SH} in \texttt{LS} back to the corresponding tokens so that tokens in \texttt{LT} effectively serves as \texttt{SH} in our transition system. 

\paragraph{\texttt{PT}} \texttt{PT} is also based on a top-down linearization, although it describes a constituent using templates: ``$p_i$ \texttt{has} $\{c_j\}$'', where $p_i$ is a constituent and $c_j$-s are its children. To describe a constituent, the indefinite article ``\textit{a}'' is used to denote a new constituent (e.g., ``... \textit{has \textbf{a} noun phrase}''). 
The definite article ``\textit{the}''\LN is used for referring to an existing constituent mentioned before (e.g., ``\textit{\textbf{the} noun phrase has} ...''), or describing a constituent whose children are all terminals (e.g., ``... \textit{has \textbf{the} noun phrase `My friend'}''). 
When describing a constituent that directly follows its mention, the determiner ``\textit{which}'' is used instead of repeating itself multiple times e.g., ``(... \textit{and the subordinating clause, \textbf{which} has} ...''). 
Sentences are joined with a semicolon ``\texttt{;}'' as the final prompt.

\subsection{Dependency Parsing (\DEP)}
\label{sec:dep-schema}

\paragraph{\texttt{LS}} \texttt{LS} uses three transitions from the arc-standard system  \cite{nivre-2004-incrementality} are used for \texttt{LS}: shift (\texttt{SH}), left arc (\texttt{<}), and right arc (\texttt{>}).

\paragraph{\texttt{LT}} \texttt{LT} for \DEP{} is obtained by replacing each \texttt{SH} in a \texttt{LS} with its corresponding token. 

\paragraph{\texttt{PT}} \texttt{PT} is derived from its \texttt{LS} sequence by removing all \texttt{SH}. Then, for each left arc creating an arc from $x_j$ to $x_i$ with dependency relation $r$ (e.g., a possessive modifier), a sentence is created by applying the template ``$x_i$ \textit{\textbf{is}} $r$ \textit{of} $x_j$''.  For each right arc creating an arc from $x_i$ to $x_j$ with the dependency relation $r$, a sentence is created with another template ``$x_i$ \textit{\textbf{has}} $r$ $x_j$''. The prompt is finalized by joining all such sentences with a semicolon.

%% file: tex/decoding-strategies.tex
\section{Decoding Strategies}
\label{sec:decoding-strategies}

To ensure well-formed output sequences that match the schemas (Section~\ref{sec:schemas}), a set of constrained decoding strategies is designed per task except for \CON, which is already tackled as S2S modeling without constrained decoding \cite{NIPS2015_277281aa, fernandez-gonzalez-gomez-rodriguez-2020-enriched}. 
Formally, given an input $\mathbf{x}$ and any partial $\mathbf{y}_{<i}$, a constrained decoding algorithm defines a  function \texttt{NextY} returning the set of all possible values for $y_i$ that can immediately follow $\mathbf{y}_{<i}$ without violating the schema. 
For brevity, subwords handling is separately explained in Section~\ref{sec:experiments}.


%
\subsection{Part-of-Speech Tagging}
\paragraph{\texttt{LS}} A finite set of \POS{} tags, $\mathcal{D}$, are collected from the training set. Specifically, \texttt{NextY} returns $\mathcal{D}$ if $i\leq n$. Otherwise, it returns \texttt{EOS}.

\paragraph{\texttt{LT}} Since tokens from the input sentence are interleaved with their \POS{} tags in \texttt{LT}, \texttt{NextY} depends on the parity of $i$, as defined in Algorithm~\ref{alg:pos-lt}.

\vspace{-0.5em}
\begin{algorithm}
  \DontPrintSemicolon
  \SetKwFunction{NextY}{NextY}
  \SetKwProg{Fn}{Function}{:}{}

  \SetKwProg{Pn}{Function}{:}{\KwRet}
  \Pn{\NextY{$\mathbf{x}, \mathbf{y}_{<i}$}}{
  
  \eIf{$i > 2n$}
{
    \KwRet\ $\{\texttt{EOS}\}$
}{
    \eIf{$i$ is even}
    {
        \KwRet\ $\{x_{\frac{i}{2}}\}$
    }{
    \KwRet\ $\mathcal{D}$
    }
}

  }  
  
 \caption{Constrained \POS-\texttt{LT}}
 \label{alg:pos-lt}
\end{algorithm}
\vspace{-1.5em}

\paragraph{\texttt{PT}} The \texttt{PT} generation can be divided into two phases: token and ``is-a-tag'' statement generations. 
A binary status $u$ is used to indicate whether $y_i$ is expected to be a token. To generate a token, an integer $k \le n$ is used to track the index of the next token. To generate an ``is-a-tag'' statement, an offset $o$ is used to track the beginning of an ``is-a-tag'' statement. Each  description $d_j$ of a \POS{} tag $t_j$ is extended to a suffix $s_j= \text{``is}\;d_j\;\text{;}$''. 

\noindent Suffixes are stored in a trie tree $\mathcal{T}$ to facilitate prefix matching between a partially generated statement and all candidate suffixes, as shown in Algorithm~\ref{alg:prefix-match}.
The full decoding is depicted in Algorithm~\ref{alg:pos-pt}.

\vspace{-0.3em}
\begin{algorithm}
  \DontPrintSemicolon
  \SetKwFunction{PrefixMatch}{PrefixMatch}
   \SetKwFunction{GetChild}{GetChild}
  \SetKwProg{Fn}{Function}{:}{}
  \SetKwProg{Pn}{Function}{:}{\KwRet}
  
    \Pn{\PrefixMatch{$\mathcal{T}, \mathbf{p}$}}{
      $\text{node} \gets \mathcal{T}$\;
      \While{$\text{node}$ and $\mathbf{p}$}{
         $\text{node} \gets node.children[p_1]$\;
         $\mathbf{p} \gets \mathbf{p}_{>1}$
      }
      \KwRet\ node
    }  
 \caption{Prefix Matching}
 \label{alg:prefix-match}
\end{algorithm}
\vspace{-0.4em}

\begin{algorithm}
  \DontPrintSemicolon
  \SetKwFunction{NextY}{NextY}
  \SetKwProg{Fn}{Function}{:}{}
  $u \gets \text{true}, k \gets 0, o \gets 0$ \\
  \SetKwProg{Pn}{Function}{:}{\KwRet}
  \Pn{\NextY{$\mathbf{x}, \mathbf{y}_{<i}$}}{
  
  \eIf{$u$}
{
    $u \gets \text{false}, k \gets k + 1, o \gets i$\\
    \KwRet\ $\{x_k\}$
}{
	$\text{node} \gets \texttt{PrefixMatch}(\mathcal{T}, \mathbf{y}_{>o})$\;
    \eIf{$\text{node}$.children is empty}
    {
    	$u \gets \text{true}$ \\
        \KwRet\ $\texttt{NextY}(\mathbf{x}, \mathbf{y}_{<i})$
    }{
    \KwRet\ $\text{node}$.children
    }
}

  }  
  
 \caption{Constrained \POS-\texttt{PT}}
 \label{alg:pos-pt}
\end{algorithm}
\vspace{-0.3em}

\subsection{Named Entity Recognition}

\paragraph{\texttt{LS}} Similar to \POS{}-\texttt{LS}, the \texttt{NextY} for \NER{} returns BIEOS tags if $i \le n$ else \texttt{EOS}.

\paragraph{\texttt{LT}} Opening tags (\texttt{<>}) in \NER{}-\texttt{LT} are grouped into a  vocabulary $\mathcal{O}$. The last generated output token $y_{i-1}$ (assuming $y_0=\texttt{BOS}$, a.k.a. beginning of a sentence) is looked up in $\mathcal{O}$ to decide what type of token will be generated next. To enforce label consistency between a pair of tags, a variable $e$ is introduced to record the expected closing tag. Reusing the definition of $k$ in Algorithm~\ref{alg:pos-pt}, decoding of \NER{}-\texttt{LT} is described in Algorithm~\ref{alg:ner-lt}.

\paragraph{\texttt{PT}} For each entity type $e_i$, its description $d_i$ is filled into the template ``is $d_i$;'' to create an ``is-a'' suffix $s_i$. Since the prompt is constructed using text while the number of entities is variable, it is not straightforward to tell whether a token belongs to an entity or an ``is-a'' suffix. Therefore, a noisy segmentation procedure is utilized to split a phrase into two parts: entity and ``is-a'' suffix. Each $s_i$ is collected into a trie $\mathcal{S}$ to perform segmentation of a partially generated phrase $\mathbf{p}$ (Algorithm~\ref{alg:segmentation}). 

\begin{algorithm}
  \DontPrintSemicolon
  \SetKwFunction{NextY}{NextY}
  \SetKwProg{Fn}{Function}{:}{}
  $k \gets 0, e \gets \text{null}$ \\
  \SetKwProg{Pn}{Function}{:}{\KwRet}
  \Pn{\NextY{$\mathbf{x}, \mathbf{y}_{<i}$}}{
  $\mathcal{Y} \gets \emptyset$ \;

  \uIf{$y_{i-1} \in \mathcal{O}$}{
    $e \gets \text{the paired closing tag of } y_{i-1}$ \;
  }
  \uElseIf{$y_{i-1}$ is $e$}{
    $e \gets \text{null}$ \;
  }
  \uElseIf{$y_{i-1} \in \mathbf{x}$}{
    $k \gets k+1$ \;
  }
  
  \uIf{$e$}{
    $\mathcal{Y} \gets \mathcal{Y} \cup \{e\}$ \;
  }
  \uElse{
	$\mathcal{Y} \gets \mathcal{Y} \cup \mathcal{O}$ \;
  }
  
  \uIf{$k > n$}
  {
      $\mathcal{Y} \gets \mathcal{Y} \cup \{\texttt{EOS}\}$ \;
  }
  \uElse{
	$\mathcal{Y} \gets \mathcal{Y} \cup \{x_k\}$ \;
  }
  
  \KwRet\ $\mathcal{Y}$
}  
  
 \caption{Constrained \NER-\texttt{LT}}
 \label{alg:ner-lt}
\end{algorithm}

\begin{algorithm}
  \DontPrintSemicolon
  \SetKwFunction{Segment}{Segment}

  \SetKwProg{Fn}{Function}{:}{}
  \SetKwProg{Pn}{Function}{:}{\KwRet}
  
    \Pn{\Segment{$\mathcal{S}, \mathbf{p}$}}{
    \For{$i \gets 1$ \KwTo $\lvert \mathbf{p} \rvert$}{
        $\text{entity}, \text{suffix} = \mathbf{p}_{\leq i}, \mathbf{p}_{>i}$ \;
        $\text{node} \gets \texttt{PrefixMatch}(\mathcal{S}, \mathbf{p}_{>i})$\;
        \uIf{$node$}
        {
            \KwRet\  \text{entity}, \text{suffix}, \text{node}
        }        
    }
      \KwRet\ \text{null}
    }  
 \caption{Segmentation}
 \label{alg:segmentation}
\end{algorithm}

\begin{algorithm}
  \DontPrintSemicolon
  \SetKwFunction{NextY}{NextY}
  \SetKwProg{Fn}{Function}{:}{}
  \SetKwProg{Pn}{Function}{:}{\KwRet}
  \Pn{\NextY{$\mathbf{x}, \mathbf{y}_{<i}$}}{
  \lIf{\texttt{EOS} in $\mathbf{y}_{<i}$}
  {
      \KwRet\ $\{\texttt{EOS}\}$
  }  
  $\mathcal{Y} \gets \emptyset$ \;
  $\mathbf{p} \gets \text{last phrase in } \mathbf{y}_{<i}$ split by ``;''\;
  \uIf{$\texttt{Segment}(\mathcal{S}, \mathbf{p})$}
  {
      \text{entity}, \text{suffix}, \text{node} $\gets \texttt{Segment}(\mathcal{S}, \mathbf{p})$ \;
      \uIf{node is terminal}
	  {
    	  $\mathcal{Y} \gets \mathcal{Y} \cup \{\texttt{EOS}\}$ \;
	  }
  	  \uElse{
	      $\mathcal{Y} \gets \mathcal{Y} \cup \text{node.children}$ \;
	      \uIf{node is root}
		  {
    		  \ForEach{occurrence $o$}{
    		      $\mathcal{Y} \gets \mathcal{Y} \cup \{x_{o+1}\}$ \;
   			  }
		  }	      
      }	  
  }
  \uElse{
	$\mathcal{Y} \gets \mathcal{Y} \cup \mathbf{x}$ \;
  }
  
  \KwRet\ $\mathcal{Y}$
}  
 \caption{Constrained \NER-\texttt{PT}}
 \label{alg:ner-pt}
\end{algorithm}

\noindent Once a segment is obtained, the decoder is constrained to generate the entity or the suffix. For the generation of an entity, string matching is used to find every occurrence $o$ of its partial generation in $\mathbf{x}$ and add the following token $x_{o+1}$ into the candidate set $\mathcal{Y}$. String matching could be noisy when an entity shares the same surface form with a non-entity phrase although no such cases are found in our datasets. Entities are generated sequentially and no nested entity is considered. To complete an ``is-a'' suffix, children of the prefix-matched node are added to the candidates (Algorithm~\ref{alg:ner-pt}).

\subsection{Constituency Parsing}

S2S on \CON{} using \texttt{LS} and \texttt{LT} has been studied and their results are good without using constrained decoding \cite{NIPS2015_277281aa, fernandez-gonzalez-gomez-rodriguez-2020-enriched}\footnote{We experimented constrained decoding with \texttt{LS} on PTB and OntoNotes, and the improvement is marginal ($+0.01$).}; thus, we focus on only \texttt{PT} in our work.

\paragraph{\texttt{PT}} The reverse linearization algorithm for \CON-\texttt{PT} is non-trivial. To restore a constituency tree from \texttt{PT}, the prompt is split into sentences creating new constituents, and sentences attaching new constituents to existing ones. 
Splitting is done by longest-prefix-matching (Algorithm~\ref{alg:longest-prefix-match}) using a trie $\mathcal{T}$ built with the definite and indefinite article versions of the description of each constituent label e.g., ``\textit{the noun phrase}'' and ``\textit{a noun phrase}'' of \texttt{NP}.

\begin{algorithm}[h]
  \DontPrintSemicolon
  \SetKwFunction{LongestPrefixMatch}{LongestPrefixMatch}
  \SetKw{Continue}{continue}
  \SetKw{Break}{break}
  \SetKwProg{Fn}{Function}{:}{}
  \SetKwProg{Pn}{Function}{:}{\KwRet}
  
    \Pn{\LongestPrefixMatch{$\mathcal{T}, \mathbf{x}$}}{
      $\text{matches} \gets \emptyset$\;
      \For{$i \gets 1$ \KwTo $\lvert \mathbf{x} \rvert+1$}{
         $\text{node} \gets \mathcal{T}.children[x_i]$\;
         \uIf{node is null}
      	  {
          	  \Continue
      	  }         
         $j \gets i+1$\;
         $v \gets \text{node.value}$\;
         \For{$k \gets j$ \KwTo $\lvert \mathbf{x} \rvert+1$}{
           $\text{node} \gets node.children[x_k]$\;
           \uIf{node is null}
      	   {
          	  \Break
      	   } 
      	   \uIf{node.value}
      	   {
          	  $v \gets \text{node.value}$\;
          	  $j \gets k+1$\;
      	   } 
		}
		\uIf{v}{
		  $\text{matches} \gets \text{matches} \cup \{(i, j, v)\}$\;
		}
      }
      \KwRet\ \text{matches}
    }  
 \caption{Longest Prefix Matching}
 \label{alg:longest-prefix-match}
\end{algorithm}

\noindent Algorithm~\ref{alg:split} describes the splitting procedure.

\vspace{-0.5em}
\begin{algorithm}
  \DontPrintSemicolon
  \SetKwFunction{Split}{Split}
  \SetKw{Continue}{continue}
  \SetKw{Break}{break}
  \SetKwProg{Fn}{Function}{:}{}
  \SetKwProg{Pn}{Function}{:}{\KwRet}
  
    \Pn{\Split{$\mathcal{T}, \mathbf{x}$}}{
      $(\text{spans}, o) \gets (\emptyset, 1)$\;
      \ForEach{$(i,j,v) \in $ \texttt{LongestPrefixMatch}($\mathcal{T}, \mathbf{x}$)}{
        \uIf{$i > o$}{
          $\text{spans} \gets \text{spans} \cup \{(o, i, \text{null})\}$\;
        }
        $\text{spans} \gets \text{spans} \cup \{(i, j, v)\}$\;
      }
      \uIf{$o < \lvert \mathbf{x} \rvert+1$}{
          $\text{spans} \gets \text{spans} \cup \{(o, \lvert \mathbf{x} \rvert+1, \text{null})\}$\;
        }
      \KwRet\ \text{spans}
    }  
 \caption{Longest Prefix Splitting }
 \label{alg:split}
\end{algorithm}
\vspace{-0.5em}

\noindent Once a prompt is split into two types of sentences, a constituency tree is then built accordingly. 
We use a variable \texttt{parent} to track the last constituent that gets attachments, and another variable \texttt{latest} to track the current new consistent that gets created.
Due to the top-down nature of linearization, the target constituent that new constituents are attached to is always among the siblings of either \texttt{parent} or the ancestors of \texttt{parent}. The search of the target constituent is described in Algorithm~\ref{alg:find-target}.

\vspace{-0.5em}
\begin{algorithm}
  \DontPrintSemicolon
  \SetKwFunction{FindTarget}{FindTarget}
  \SetKwProg{Fn}{Function}{:}{}
  \SetKwProg{Pn}{Function}{:}{\KwRet}
  
    \Pn{\FindTarget{$\text{parent}, \text{label}$}}{
      \While{$\text{parent}$}{
         \ForEach{sibling of parent}{
          \uIf{sibling.label is label and sibling has no children}{
              \KwRet\ \text{sibling}
           }
         }
         $\text{parent} \gets \text{parent.parent}$
      }
      
      \KwRet\ null
    }  
 \caption{Find Target}
 \label{alg:find-target}
\end{algorithm}
\vspace{-0.5em}

\noindent Algorithm~\ref{alg:con-pt} shows the final reverse linearization.

\begin{algorithm}
  \DontPrintSemicolon
  \SetKwFunction{Reverse}{Reverse}
  \SetKw{Continue}{continue}
  \SetKw{Break}{break}
  \SetKwProg{Fn}{Function}{:}{}
  \SetKwProg{Pn}{Function}{:}{\KwRet}
  
    \Pn{\Reverse{$\mathcal{T}, \mathbf{x}$}}{
      $\text{root} \gets \text{parent} 
      \gets \text{new TOP-tree}$\;
      $\text{latest} \gets \text{null}$ \;
      \ForEach{$(i,j,v) \in $ \texttt{Split}($\mathcal{T}, \mathbf{x}$)}{
        \uIf{$v$}{
            \uIf{$\mathbf{x}_{i:j}$ starts with ``the''}{
              target $\gets$ \texttt{FindTarget}(parent, $v$)\;
            }
            \uElse{
              latest $\gets$ new $v$-tree \;
              add latest to parent.children \;
              latest.parent $\gets$ parent \;
            }  
        }
        \uElse{
          \uIf{$\mathbf{x}_{i:j}$ starts with ``has'' or ``which has''}{
            parent $\gets$ latest \;
          }
          add tokens in ``'' into latest\;
        }  
      }

      \KwRet\ \text{root}
    }  
 \caption{Reverse \CON-\texttt{PT} }
 \label{alg:con-pt}
\end{algorithm}

\subsection{Dependency Parsing}

\paragraph{\texttt{LS}} Arc-standard \cite{nivre-2004-incrementality} transitions are added to a candidate set and only transitions permitted by the current parsing state are allowed.

\paragraph{\texttt{LT}} \DEP{}-\texttt{LS} replaces all \texttt{SH} transitions with input tokens in left-to-right order. Therefore, an incremental offset is kept to generate the next token in place of each \texttt{SH} in \DEP{}-\texttt{LT}.

\paragraph{\texttt{PT}} \DEP{}-\texttt{PT} is more complicated than \CON{}-\texttt{PT} because each sentence contains one more token. Its generation is therefore divided into 4 possible states: first token (\texttt{1st}), relation (\texttt{rel}), second token (\texttt{2ed}), and semicolon. 
An arc-standard transition system is executed synchronously with constrained decoding since \texttt{PT} is essentially a simplified transition sequence with all \texttt{SH} removed.
Let $\mathbf{b}$ and $\mathbf{s}$ be the system buffer and stack, respectively.
Let $\mathbf{c}$ be a set of candidate tokens that will be generated in $\mathbf{y}$, which initially contains all input tokens and an inserted token ``sentence'' that is only used to represent the root in ``the sentence has a root $\ldots$'' 
A token is removed from $\mathbf{c}$ once it gets popped out of $\mathbf{s}$.
Since \DEP{}-\texttt{PT} generates no \texttt{SH}, each input token $x_j$ in $\mathbf{y}$ effectively introduces \texttt{SH}(s) till it is pushed onto $\mathbf{s}$ at index $i$ ($i \in \{1, 2\}$), as formally described in Algorithm~\ref{alg:recall-shift}.

\begin{algorithm}
  \DontPrintSemicolon
  \SetKwFunction{RecallShift}{RecallShift}
  \SetKwProg{Fn}{Function}{:}{}
  \SetKwProg{Pn}{Function}{:}{\KwRet}
  
    \Pn{\RecallShift{$\text{system}, \text{i}, x_j$}}{
      \While{$\text{system}.s_i$ is not $x_j$}{
        system.apply(\texttt{SH}) \;
      }
    }  
 \caption{Recall Shift}
 \label{alg:recall-shift}
\end{algorithm}

\noindent 
After the first token is generated, its offset $o$ in $\mathbf{y}$ is recorded such that the following relation sequence $\mathbf{y}_{i>o}$ can be located. 
To decide the next token of $\mathbf{y}_{i>o}$, it is then prefix-matched with a trie $\mathcal{T}$ built with the set of ``has-'' and ``is-'' dependency relations. 
The children of the prefix-matched node are considered candidates if it has any. 
Otherwise, the dependency relation is marked as completed.
Once a relation is generated, the second token will be generated in a similar way. 
Finally, upon the completion of a sentence, the transition it describes is applied to the system and $\mathbf{c}$ is updated accordingly. 
The full procedure is described in Algorithm~\ref{alg:dep-pt}. 
Since a transition system has been synchronously maintained with constrained decoding, no extra reverse linearization is needed.

\begin{algorithm}[!]
  \DontPrintSemicolon
  \SetKwFunction{NextY}{NextY}
  \SetKwProg{Fn}{Function}{:}{}
  \SetKwProg{Pn}{Function}{:}{\KwRet}
  \resizebox{0.95\columnwidth}{!}{$(\text{status}, \text{transition}, t_1, t_2, o) \gets (\text{1st}, \text{null}, \text{null}, \text{null}, 0)$}
  $\mathbf{c} \gets \{\texttt{sentence}\} \cup \mathbf{y}$\;
  \Pn{\NextY{$\mathbf{x}, \mathbf{y}_{<i}$}}{
  \uIf{\texttt{EOS} in $\mathbf{y}_{<i}$}
  {
      \KwRet\ $\{\texttt{EOS}\}$ \;
  }  
  $\mathcal{Y} \gets \emptyset$ \;
  $\mathbf{p} \gets \text{last phrase in } \mathbf{y}_{<i}$ split by ``;''\;
  \uIf{status is semicolon}
  {
    $\mathcal{Y} \gets \mathcal{Y} \cup \{\text{;}\}$ \;
    $\text{status} \gets \text{1st}$\;
  }
  \uElseIf{status is 1st}{
     $\mathcal{Y} \gets \mathcal{Y} \cup \mathbf{c}$ \;
     $o \gets i$ \;
     $\text{status} \gets \text{rel}$ \;
  }
  \uElseIf{status is rel}{
     $t_1 \gets y_{o}$\;
     $\text{node} \gets \texttt{PrefixMatch}(\mathcal{T}, \mathbf{y}_{>o})$\;
     \uIf{node.children} {
      $\mathcal{Y} \gets \mathcal{Y} \cup \{\text{node.children}\}$ \;
    }
    \uElse{
      $relation \gets$ the relation in $\mathbf{y}_{>o}$ \;
      \uIf{$\mathbf{y}_{>o}$ starts with ``is''} {
        $transition \gets$ LA-relation \;
      }
      \uElse{
        $transition \gets$ RA-relation \;
      }    
      $\text{status} \gets \text{2ed}$ \;  
    }
  }
  \uElseIf{status is 2ed}{
     $\mathcal{Y} \gets \mathcal{Y} \cup \mathbf{c}$ \;
     $\text{status} \gets \text{semicolon}$ \;
  }  
  \uElseIf{status is semicolon}{
     $t_2 \gets y_{i-1}$\;    
     $\mathcal{Y} \gets \mathcal{Y} \cup \{\text{;}\}$ \;
     $\texttt{RecallShift}(\text{system}, 1, t_2)$ \;
     $\texttt{RecallShift}(\text{system}, 2, t_1)$ \;
     \uIf{transition starts with LA} {
       remove $s_1$ from $\mathbf{c}$ \;
     }
     \uElse{
       remove $s_2$ from $\mathbf{c}$ \;
     }
     $\text{system}.$\texttt{apply}$(\text{transition})$\;
     \uIf{system is terminal} {
       $\mathcal{Y} \gets \mathcal{Y} \cup \{\texttt{EOS}\}$ \;
     }
     $\text{status} \gets \text{1st}$\;
  }  
  
  \KwRet\ $\mathcal{Y}$
}  
  
 \caption{Constrained \DEP-\texttt{PT}}
 \label{alg:dep-pt}
\end{algorithm}

%% file: tex/experiments.tex
\section{Experiments}
\label{sec:experiments}

For all tasks, BART-Large \cite{lewis-etal-2020-bart} is finetuned as our underlying S2S model. 
We also tried T5 \cite{JMLR:v21:20-074} although its performance was less satisfactory. 
Every model is trained three times using different random seeds and their average scores and standard deviations on the test sets are reported.
Our models are experimented on the OntoNotes 5 \cite{weischedel2013ontonotes} using the data split suggested by \citet{pradhan-etal-2013-towards}. 
In addition, two other popular datasets are used for fair comparisons to previous works: the Wall Street Journal corpus from the Penn Treebank 3~\cite{Marcus1993} for \POS{}, \DEP{} and \CON, as well as the English portion of the CoNLL'03 dataset \cite{tjong-kim-sang-de-meulder-2003-introduction} for \NER{}.

Each token is independently tokenized using the subword tokenizer of BART and merged into an input sequence. The boundary information for each token is recorded to ensure full tokens are generated in \texttt{LT} and \texttt{PT} without broken pieces. To fit in the positional embeddings of BART, sentences longer than 1,024 subwords are discarded, which include 1 sentence from the Penn Treebank 3 training set, and 24 sentences from the OntoNotes 5 training set. Development sets and test sets are not affected.

\subsection{Part-of-Speech Tagging}

Token level accuracy is used as the metric for \POS{}. 
\texttt{LT} outperforms \texttt{LS} although \texttt{LT} is twice as long as \texttt{LS}, suggesting that textual tokens positively impact the learning of the decoder (Table~\ref{tbl:pos}). 
\texttt{PT} performs almost the same with \texttt{LT}, perhaps due to the fact that \POS{} is not a task requiring a powerful decoder.

\begin{table}[htbp!]
\centering\resizebox{\columnwidth}{!}{
\begin{tabular}{c|c|c}
\toprule
Model & PTB        & OntoNotes            \\
\midrule
\citet{bohnet:18a} & 97.96        & -            \\
\citet{he-choi-2021-stem} & - & 98.32 $\pm$ 0.02\\
\midrule
\texttt{LS}          & 97.51 $\pm$ 0.11 & 98.21 $\pm$ 0.02 \\
\texttt{LT}          & \textbf{97.70} $\pm$ 0.02 & \textbf{98.40} $\pm$ 0.01 \\
\texttt{PT}          & 97.64 $\pm$ 0.01 & 98.37 $\pm$ 0.02 \\
\bottomrule
\end{tabular}}
  \caption{Results for \POS.}
  \label{tbl:pos}
\vspace{-0.5em}
\end{table}

\subsection{Named Entity Recognition}

For CoNLL'03, the provided splits without merging the development and training sets are used. For OntoNotes 5, the same splits as \citet{chiu:16a, li-etal-2017-leveraging, ghaddar-langlais-2018-robust, he-choi-2019, he-choi-2021-stem} are used. 
Labeled span-level F$_1$ score is used for evaluation. 

\noindent We acknowledge that the performance of NER systems can be largely improved by rich embeddings \cite{wang-etal-2021-automated}, document context feature \cite{yu-etal-2020-named}, dependency tree feature \cite{xu-etal-2021-better}, and other external resources. While our focus is the potential of S2S, we mainly consider two strong baselines that also use BART as the only external resource: the generative BART-Pointer framework \cite{yan-etal-2021-unified-generative} and the recent template-based BART NER \cite{cui-etal-2021-template}.

\begin{table}[htbp!]
\centering\resizebox{\columnwidth}{!}{
\begin{tabular}{c|c|c}
\toprule
     Model                &  CoNLL'03 & OntoNotes 5          \\
\midrule
  \citet{clark-etal-2018-semi}                           & 92.60 & -       \\
  \citet{peters-etal-2018-deep} & 92.22 & -  \\
  \citet{akbik-etal-2019-pooled} & 93.18 & -  \\
  \citet{strakova-etal-2019-neural}         & 93.07 & -      \\
  \citet{yamada-etal-2020-luke}        & 92.40  & -    \\
  \citet{yu-etal-2020-named}$\dagger$              & 92.50 &   89.83     \\
  \citet{yan-etal-2021-unified-generative}$\ddagger^\mathcal{S}$      & {93.24}  & {90.38}   \\
  \citet{cui-etal-2021-template}$^\mathcal{S}$        & 92.55  & -      \\
  \citet{he-choi-2021-stem} & -  & 89.04 $\pm$ 0.14      \\
  \citet{wang-etal-2021-automated} & 94.6  & -      \\
  \citet{zhu-li-2022-boundary} & -  & 	91.74	      \\
  \citet{ye-etal-2022-packed} & -  & 	91.9	      \\
\midrule
\texttt{LS}  & 70.29 $\pm$ 0.70  & 84.61 $\pm$ 1.18\\
\texttt{LT}  & 92.75 $\pm$ 0.03 & 89.60 $\pm$ 0.06\\
\texttt{PT}  & \textbf{93.18} $\pm$ 0.04 & \textbf{90.33} $\pm$ 0.04\\
\bottomrule
\end{tabular}}
     \caption{Results for \NER. $\mathcal{S}$ denotes S2S.}
     \label{tbl:ner}
\vspace{-0.5em}
\end{table}

\noindent As shown in Table~\ref{tbl:ner}, \texttt{LS} performs the worst on both datasets, possibly attributed to that the autoregressive decoder overfits the high-order left-to-right dependencies of BIEOS tags. 
\texttt{LT} performs close to the BERT-Large biaffine model \cite{yu-etal-2020-named}. \texttt{PT} performs comparably well with the Pointer Networks approach \cite{yu-etal-2020-named} and it outperforms the template prompting \cite{cui-etal-2021-template} by a large margin, suggesting S2S has the potential to learn structures without using external modules.

\subsection{Constituency Parsing}

\label{sec:exp-con}

All \POS{} tags are removed and not used in training or evaluation. Terminals belonging to the same non-terminal are flattened into one constituent before training and unflattened in post-processing. The standard constituent-level F-score produced by the EVALB \footnote{\url{https://nlp.cs.nyu.edu/evalb/}} is used as the evaluation metric.


Table~\ref{tbl:con} shows the results on OntoNotes 5 and PTB 3. Incorporating textual tokens into the output sequence is important on OntoNotes 5, leading to a +0.9 F-score, while it is not the case on PTB 3. It is possibly due to that OntoNotes is more diverse in domains, requiring a higher utilization of pre-trained S2S for domain transfer. \texttt{PT} performs the best, and it has a competitive performance to recent works, despite that it uses no extra decoders.

\begin{table}[htbp!]
\centering\resizebox{\columnwidth}{!}{
  \begin{tabular}{c|c|c}
  \toprule
  Model & PTB 3       & OntoNotes 5           \\
  \midrule
  \citet{fernandez-gonzalez-gomez-rodriguez-2020-enriched}$^\mathcal{S}$ & 91.6         &       -       \\
  \citet{mrini-etal-2020-rethinking}    & 96.38        &      -        \\
  \citet{he-choi-2021-stem} & - & 94.43 $\pm$ 0.03\\
  \citet{yang-tu-2022-bottom}$^\mathcal{S}$ & 96.01 & - \\
  \midrule
  \texttt{LS}      & 95.23 $\pm$ 0.08 & 93.40 $\pm$ 0.31 \\
  \texttt{LT}      & 95.24 $\pm$ 0.04 & 94.32 $\pm$ 0.11 \\
  \texttt{PT}      & \textbf{95.34} $\pm$ 0.06 & \textbf{94.55} $\pm$ 0.03 \\
  \bottomrule
  \end{tabular}}
    \caption{Results for \CON. $\mathcal{S}$ denotes S2S}
    \label{tbl:con}
\vspace{-1.5em}
  \end{table}

\subsection{Dependency Parsing}

The constituency trees from PTB and OntoNotes are converted into the Stanford dependencies v3.3.0 \cite{de2008stanford} for \DEP{} experiments. 40 and 1 non-projective trees are removed from the training and development sets of PTB 3, respectively. For OntoNotes 5, these numbers are 262 and 28. Test sets are not affected.

\begin{table}[htbp!]
  \begin{subtable}[h]{\columnwidth}
\centering\resizebox{\columnwidth}{!}{

\begin{tabular}{c|c|c}
\toprule
     Model              & UAS          & LAS          \\
\midrule
\citet{wiseman-rush-2016-sequence}$^\mathcal{S}$ & 91.17 & 87.41\\
\citet{zhang-etal-2017-stack}$^\mathcal{S}$ & 93.71 & 91.60\\
\citet{li-etal-2018-seq2seq}$^\mathcal{S}$ & 94.11 & 92.08\\
\citet{mrini-etal-2020-rethinking} & 97.42        & 96.26        \\
\midrule
\texttt{LS}                   & 92.83 $\pm$ 0.43 & 90.50 $\pm$ 0.53 \\
\texttt{LT}                   & 95.79 $\pm$ 0.07 & 93.17 $\pm$ 0.16 \\
\texttt{PT}                   & \textbf{95.91} $\pm$ 0.06 & \textbf{94.31} $\pm$ 0.09 \\
\bottomrule
\end{tabular}}
\caption{PTB results for \DEP.}
\label{tbl:dep-ptb}
  \end{subtable}
  
  
\begin{subtable}[h]{\columnwidth}
\centering\resizebox{\columnwidth}{!}{

\begin{tabular}{c|c|c}
\toprule
         Model          & UAS          & LAS          \\
\midrule
\citet{he-choi-2021-stem} & 95.92 $\pm$ 0.02        & 94.24 $\pm$ 0.03        \\
\midrule
\texttt{LS} & 86.54 $\pm$ 0.12 & 83.84 $\pm$ 0.13 \\
\texttt{LT} & 94.15 $\pm$ 0.14 & 91.27 $\pm$ 0.19 \\
\texttt{PT} & \textbf{94.51} $\pm$ 0.22 &	\textbf{92.81} $\pm$ 0.21\\
\bottomrule
\end{tabular}}
\caption{OntoNotes results for \DEP.}
\label{tbl:dep-ontonotes}
   \end{subtable}
   \caption{Results for \DEP. $\mathcal{S}$ denotes S2S.}
   \label{tbl:dep}
\vspace{-0.5em}
\end{table}

\noindent As shown in Table~\ref{tbl:dep}, textual tokens are crucial in learning arc-standard transitions using S2S, leading to +2.6 and +7.4 LAS improvements, respectively. Although our \texttt{PT} method underperforms recent state-of-the-art methods, it has the strongest performance among all S2S approaches. Interestingly, our S2S model manages to learn a transition system without explicitly modeling the stack, the buffer, the partial parse, or pointers.

\noindent We believe that the performance of \DEP{} with S2S can be further improved with a larger and more recent pretrained S2S model and dynamic oracle \cite{goldberg-nivre-2012-dynamic}.

%% file: tex/analysis.tex
\section{Analysis}
\label{sec:analysis}

\subsection{Ablation Study}
\label{sec:abulation}

We perform an ablation study to show the performance gain of our proposed constrained decoding algorithms on different tasks. Constrained decoding algorithms (\texttt{CD}) are compared against free generation (\texttt{w/o CD}) where a model freely generates an output sequence that is later post-processed into task-specific structures using string-matching rules. Invalid outputs are patched to the greatest extent, e.g., \POS{} label sequences are padded or truncated.

\begin{table}[htbp!]
    \begin{subtable}[h]{\columnwidth}
  \centering\small{
    \begin{tabular}{c|c|c}
    \toprule
    Model & PTB        & OntoNotes            \\
    \midrule
    \texttt{LS}          & 97.51 $\pm$ 0.11 & 98.21 $\pm$ 0.02 \\
    \texttt{w/o CD}      & 97.51 $\pm$ 0.11&98.21 $\pm$ 0.02 \\
    \midrule
    \texttt{LT}          & 97.70 $\pm$ 0.02 & 98.40 $\pm$ 0.01 \\
    \texttt{w/o CD}      & 97.67 $\pm$ 0.02&98.39 $\pm$ 0.01 \\
    \midrule
    \texttt{PT}          & 97.64 $\pm$ 0.01 & 98.37 $\pm$ 0.02 \\
    \texttt{w/o CD}      & 97.55 $\pm$ 0.02&98.29 $\pm$ 0.05 \\
    \bottomrule
    \end{tabular}}
    \caption{Accuracy of ablation tests for \POS.}
    \label{tbl:pos-ablation}
    \end{subtable}
    
  \begin{subtable}[h]{\columnwidth}
  \centering\small
  \begin{tabular}{c|c|c}
    \toprule
         Model                &  CoNLL 03 & OntoNotes 5          \\
    \midrule
    \texttt{LS}  & 70.29 $\pm$ 0.70  & 84.61 $\pm$ 1.18\\
    \texttt{w/o CD}      & 66.33 $\pm$ 0.73 & 84.57 $\pm$ 1.16 \\
    \midrule
    \texttt{LT}  & 92.75 $\pm$ 0.03 & 89.60 $\pm$ 0.06\\
    \texttt{w/o CD}      & 92.72 $\pm$ 0.02 & 89.50 $\pm$ 0.07 \\
    \midrule
    \texttt{PT}  & 93.18 $\pm$ 0.04 & 90.33 $\pm$ 0.04\\
    \texttt{w/o CD}      & 93.12 $\pm$ 0.06 & 90.23 $\pm$ 0.05 \\
    \bottomrule
    \end{tabular}
  \caption{F1 of ablation tests for \NER.}
  \label{tbl:ner-ablation}
     \end{subtable}

     \begin{subtable}[h]{\columnwidth}
        \centering\small{
          \begin{tabular}{c|c|c}
          \toprule
          Model & PTB        & OntoNotes            \\
          \midrule
          \texttt{LS}          & 90.50 $\pm$ 0.53  & 83.84 $\pm$ 0.13 \\
          \texttt{w/o CD}      & 90.45 $\pm$ 0.47  & 83.78 $\pm$ 0.13 \\
          \midrule
          \texttt{LT}          & 93.17 $\pm$ 0.16  & 91.27 $\pm$ 0.19 \\
          \texttt{w/o CD}      & 93.12 $\pm$ 0.14  & 91.05 $\pm$ 0.20 \\
          \midrule
          \texttt{PT}          & 94.31 $\pm$ 0.09 & 92.81 $\pm$ 0.21 \\
          \texttt{w/o CD}      & 81.50 $\pm$ 0.27 & 81.76 $\pm$ 0.36 \\
          \bottomrule
          \end{tabular}}
          \caption{LAS of ablation tests for \DEP.}
          \label{tbl:dep-ablation}
    \end{subtable}


     \caption{Ablation test results.}
     \label{tbl:ablation}
  \end{table}

\noindent As shown in Table~\ref{tbl:ablation}, ablation of constrained decoding seldom impacts the performance of \texttt{LS} on all tasks, suggesting that the decoder of seq2seq can acclimatize to the newly added label tokens. Interestingly, the less performant \NER-\texttt{LS} model degrades the most, promoting the necessity of constrained decoding for weaker seq2seq models.
The performance of \texttt{LT} on all tasks is marginally degraded when constrained decoding is ablated, indicating the decoder begins to generate structurally invalid outputs when textual tokens are freely generated. This type of problem seems to be exacerbated when more tokens are freely generated in the \texttt{PT} schemas, especially for the \DEP-\texttt{PT}.

Unlike \POS{} and \NER{}, \DEP{} is more prone to hallucinated textual tokens as early errors in the transition sequence get accumulated in the arc-standard system which shifts all later predictions off the track. It is not yet a critical problem as \texttt{LS} generates no textual tokens while a textual token in \texttt{LT} still serves as a valid shift action even if it is hallucinated. However, a hallucinated textual token in \texttt{PT} is catastrophic as it could be part of any arc-standard transitions. As no explicit shift transition is designed, a hallucinated token could lead to multiple instances of missing shifts in Algorithm~\ref{alg:dep-pt}.

\subsection{Case Study}

To facilitate understanding and comparison of different models, a concrete example of input (\textbf{I}), gold annotation (\textbf{G}) and actual model prediction per each schema is provided below for each task. Wrong predictions and corresponding ground truth are highlighted in \textcolor{red}{red} and \textcolor{teal}{teal} respectively.

\paragraph{\POS} In the following example, only \texttt{PT} correctly detects the past tense (\texttt{VBD}) of ``put''.

\vspace{0.5em}
\noindent\resizebox{\columnwidth}{!}{\begin{tabular}{@{}r@{\hspace{0.3em}}l@{}}
\textbf{I:} & The word I put in boldface is extremely interesting. \\
\textbf{G:} & \texttt{DT NN PRP \textcolor{teal}{VBD} IN NN VBZ RB \textcolor{teal}{JJ .}} \\
\textbf{\texttt{LS}:} & \texttt{DT NN PRP \textcolor{red}{VBP} IN NN VBZ RB \textcolor{red}{RB JJ}} \\
\textbf{\texttt{LT}:}
 & The\texttt{/DT} word\texttt{/NN} I\texttt{/PRP} put\texttt{/\textcolor{red}{VBP}} in\texttt{/IN} \\
 & boldface\texttt{/NN} is\texttt{/VBZ} extr.\texttt{/RB} interesting\texttt{/JJ} .\texttt{/.}\\
\textbf{\texttt{PT}:}
 & ``The'' is a determiner; ``word'' is a singular noun;\\
 & ``I'' is a personal pronoun; \textcolor{teal}{``put'' is a past tense verb;}\\
 & ``in'' is a preposition or subordinating conjunction;\\
 & ``boldface'' is a singular noun; ``is'' is a 3rd person \\
 & singular present verb; ``extremely'' is an adverb;\\
 & ``interesting'' is an adjective; ``.'' is a period.
\end{tabular}}

%
%
%

\paragraph{\NER} In the following example, \texttt{LS} and \texttt{LT} could not correctly recognize ``HIStory'' as an art work possibly due to its leading uppercase letters.

\vspace{0.5em}
\noindent\resizebox{\columnwidth}{!}{\begin{tabular}{@{}r@{\hspace{0.3em}}l@{}}
\textbf{I:} & Large image of the Michael Jackson HIStory statue. \\
\textbf{G:} & $
\text{Large image of the}\,\underbrace{\text{Michael Jackson}}_{\texttt{PERSON} (\texttt{PER})} \underbrace{\text{HIStory}}_{\textcolor{teal}{\texttt{WOA}}}\,\text{statue.}
$\\
\end{tabular}}

\noindent\resizebox{\columnwidth}{!}{\begin{tabular}{@{}r@{\hspace{0.3em}}l@{}}
\textbf{\texttt{LS}:} & \texttt{O O O O B-PER E-PER \textcolor{red}{S-ORG} O O O} \\
\textbf{\texttt{LT}:}
 & Large image of the \texttt{<PERSON>}Michael Jackson \\
 & \texttt{</PERSON>} \textcolor{red}{HIStory} statue. \\
\textbf{\texttt{PT}:}
 & ``Michael Jackson'' is a person; \\
 & \textcolor{teal}{``HIStory'' is an art work}.\\
\end{tabular}}

%
%
%
%

\paragraph{\CON} As highlighted with strikeout text below, \texttt{LS} and \texttt{LT} failed to parse ``how much'' as a wh-noun phrase and a wh-adverb phrase respectively.

\vspace{0.5em}
\noindent\resizebox{\columnwidth}{!}{\begin{tabular}{@{}r@{\hspace{0.3em}}l@{}}
\textbf{I:} & It's crazy how much he eats. \\
\textbf{G:}
 & \texttt{(S} \texttt{(NP} \texttt{(NP} It\texttt{))} \texttt{(VP} 's \texttt{(ADJP} crazy\texttt{)} \texttt{(SBAR} \\
 & \textcolor{teal}{\texttt{(WHNP} \texttt{(WHADJP}} how much\textcolor{teal}{\texttt{))}} \texttt{(S} \texttt{(NP} he\texttt{)} \\
 & \texttt{(VP} eats\texttt{))))} .\texttt{)}\\
\textbf{\texttt{LS}:}
 & \texttt{N-S} \texttt{N-NP} \texttt{N-NP} \texttt{SH} \texttt{RE} \texttt{RE} \texttt{N-VP} \texttt{SH} \texttt{N-ADJP} \texttt{SH} \\
 & \texttt{RE} \texttt{N-SBAR} \textcolor{red}{\st{\texttt{N-WHNP}}} \texttt{N-WHADVP} \texttt{SH} \texttt{SH} \texttt{RE} \textcolor{red}{\st{\texttt{RE}}} \\
 & \texttt{N-S} \texttt{N-NP} \texttt{SH} \texttt{RE} \texttt{N-VP} \texttt{SH} \texttt{RE} \texttt{RE} \texttt{RE} \texttt{RE} \texttt{SH} \texttt{RE} \\
\textbf{\texttt{LT}:}
 & \texttt{(S} \texttt{(NP} \texttt{(NP} It\texttt{))} \texttt{(VP} 's \texttt{(ADJP} crazy\texttt{)} \texttt{(SBAR}\\
 & \texttt{(WHNP} \textcolor{red}{\st{(\texttt{WHADJP}}} how much\textcolor{red}{\st{\texttt{)}}}) \texttt{(S} \texttt{(NP} he\texttt{)}\\
 & \texttt{(VP} eats\texttt{))))} .\texttt{)}\\
\textbf{\texttt{PT}:}
 & a sentence has a simple clause, which has a noun\\
 & phrase and a verb phrase and ``.''; the noun phrase\\
 & has a noun phrase ``It'', the verb phrase has ``'s''\\
 & and an adjective phrase ``crazy'' and a subordinating\\
 & clause, which has a wh-noun phrase and a simple\\
 & clause; \textcolor{teal}{the wh-noun phrase has a wh-adjective}\\
 & \textcolor{teal}{phrase ``how much''}, the simple clause has a noun\\
 & phrase ``he'' and a verb phrase ``eats''.\\
\end{tabular}}

%
%
%
%

\paragraph{\DEP} In the following example, \texttt{LS} incorrectly attached ``so out of'' to ``place'', and \texttt{LT} wrongly attached ``so'' to ``looks''.

\vspace{0.5em}
\noindent{\small\begin{tabular}{@{}r@{\hspace{0.3em}}l@{}}
\textbf{I:} & It looks so out of place. \\
\textbf{G:} & \includegraphics[scale=1.25]{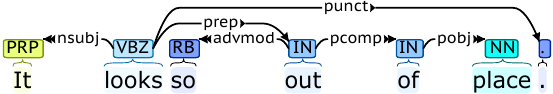} \\
\textbf{\texttt{LS}:}
 & \texttt{SH} \texttt{SH} \texttt{LA-nsubj} \texttt{SH} \texttt{SH} \textcolor{red}{\texttt{SH}} \texttt{SH} \texttt{LA-advmod}\\
 & \texttt{LA-advmod} \texttt{LA-advmod} \texttt{RA-acomp} \texttt{SH}\\
 & \texttt{RA-punct} \texttt{RA-root}\\
\textbf{\texttt{LT}:}
 & It looks \texttt{LA-nsubj} so out \textcolor{red}{of} place \texttt{RA-pobj}\\
 & \texttt{RA-pcomp} \texttt{RA-prep} \texttt{RA-ccomp} . \texttt{RA-punct}\\
 & \texttt{RA-root} \\
\textbf{\texttt{PT}:}
 & ``It'' is a nominal subject of ``looks''; \textcolor{teal}{``so'' is an}\\
 & \textcolor{teal}{adverbial modifier of ``out''}; ``of'' has an object of\\
 & a preposition ``place''; ``out'' has a prepositional\\
 & complement ``of''; ``looks'' has a prepositional\\
 & modifier ``out''; ``looks'' has a punctuation ``.'';\\
 & ``sentence'' has a root ``looks''. \\
\end{tabular}}

%
%
%
%

\subsection{Design Choices}
\label{sec:design_choices}

In the interest of experimentally comparing the schema variants, we would like each design we consider to be equivalent in some systematic way. To this end, we fix other aspects and variate two dimensions of the prompt design, lexicality and verbosity, to isolate the impact of individual variables.

\paragraph{Lexicality} We call the portion of textual tokens in a sequence its lexicality. Thus, \texttt{LS} and \texttt{PT} have zero and full lexicality respectively, while \texttt{LT} falls in the middle. To tease apart the impact of lexicality, we substitute the lexical phrases with corresponding tag abbreviations in \texttt{PT} on \POS{} and \DEP, e.g., ``friend'' is a \textit{noun} $\rightarrow$ ``friend'' is a \texttt{NN}, ``friend'' is a \textit{nominal subject} of ``bought'' $\rightarrow$ ``friend'' is a \texttt{nsubj} of ``bought''. 
Tags are added to the BART vocabulary and learned from scratch as \texttt{LS} and \texttt{LT}.

\begin{table}[htbp!]
  \centering\small
  \begin{tabular}{c|c|c}
  \toprule
  Model & PTB 3       & OntoNotes 5           \\
  \midrule
  \texttt{POS-PT}          & \textbf{97.64} $\pm$ 0.01 & \textbf{98.37} $\pm$ 0.02 \\
  \texttt{dec.LEX}         & 97.63 $\pm$ 0.02 & 98.35 $\pm$ 0.03 \\
  \midrule
  \texttt{DEP-PT}       & \textbf{94.31} $\pm$ 0.09 & \textbf{92.81} $\pm$ 0.21 \\
  \texttt{dec.LEX}      & 93.89 $\pm$ 0.18 & 91.19 $\pm$ 0.86 \\
  \bottomrule
  \end{tabular}
    \caption{Study of lexicality on \POS{} and \DEP.}
    \label{tbl:lexicality}
\vspace{-0.5em}
  \end{table}

\noindent As shown in Table~\ref{tbl:lexicality}, decreasing the lexicality of \texttt{PT} marginally degrades the performance of S2S on \POS{}. On \DEP, the performance drop is rather significant. Similar trends are observed comparing \texttt{LT} and \texttt{LS} in Section~\ref{sec:experiments}, confirming that lexicons play an important role in prompt design.

\begin{table}[htbp!]
  \centering\small
  \begin{tabular}{c|c|c}
  \toprule
       Model                &  CoNLL 03 & OntoNotes 5          \\
  \midrule
  \texttt{NER-PT}  & \textbf{93.18} $\pm$ 0.04 & \textbf{90.33} $\pm$ 0.04 \\
  \texttt{inc.VRB}  & 92.47 $\pm$ 0.03 & 89.63 $\pm$ 0.23 \\
  \toprule
  Model                &  PTB 3 & OntoNotes 5          \\
\midrule
\texttt{CON-PT}  & \textbf{95.34} $\pm$ 0.06 & \textbf{94.55} $\pm$ 0.03 \\
\texttt{inc.VRB}  & 95.19 $\pm$ 0.06 & 94.02 $\pm$ 0.49 \\
  \bottomrule
  \end{tabular}
       \caption{Study of verbosity on \NER{} and \CON.}
       \label{tbl:verbosity}
\vspace{-0.5em}
  \end{table}

\paragraph{Verbosity} Our \texttt{PT} schemas on \NER{} and \CON{} are designed to be as concise as human narrative, and as easy for S2S to generate. Another design choice would be as verbose as some \texttt{LS} and \texttt{LT} schemas. To explore this dimension, we increase the verbosity of \NER-\texttt{PT} and \CON-\texttt{PT} by adding ``isn't an entity'' for all non-entity tokens and substituting each ``which'' to its actual referred phrase respectively. The results are presented in Table~\ref{tbl:verbosity}. Though increased verbosity would eliminate any ambiguity, unfortunately, it hurts the performance. Emphasizing a token ``isn't an entity'' might encounter the over-confidence issue as the boundary annotation might be ambiguous in gold NER data \cite{zhu-li-2022-boundary}. \texttt{CON-PT} deviates from human language style when reference is forbidden, which eventually makes it lengthy and hard to learn.

\begin{figure*}[ht]
  \centering
  \begin{subfigure}[b]{0.24\textwidth}
      \centering
      \includegraphics[width=\textwidth]{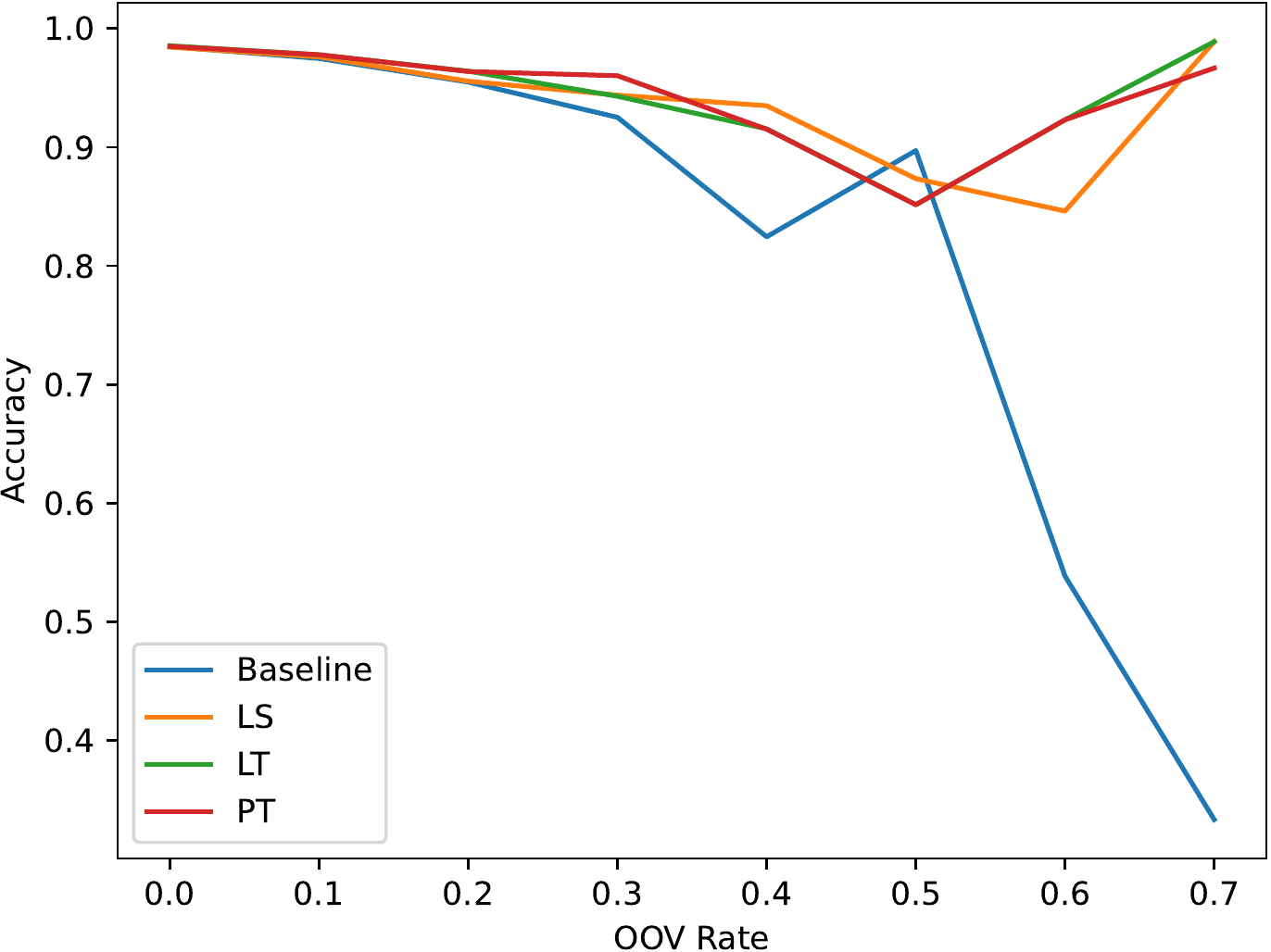}
      \caption{OOV for \POS}
      \label{fig:pos-oov}
  \end{subfigure}
  \hfill
  \begin{subfigure}[b]{0.24\textwidth}
      \centering
      \includegraphics[width=\textwidth]{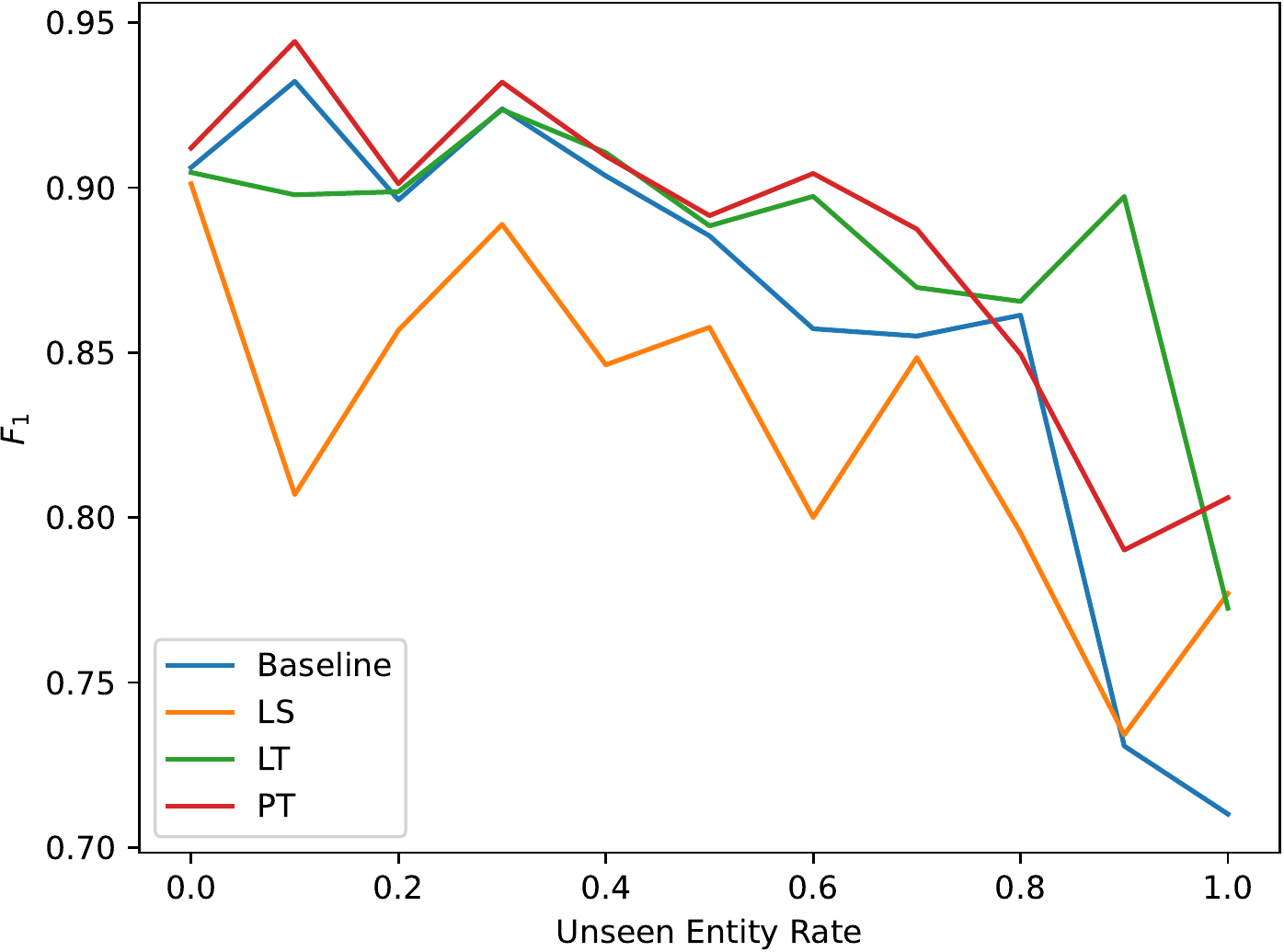}
      \caption{Unseen entities for \NER}
      \label{fig:ner-oov}
  \end{subfigure}
  \hfill
  \begin{subfigure}[b]{0.24\textwidth}
      \centering
      \includegraphics[width=\textwidth]{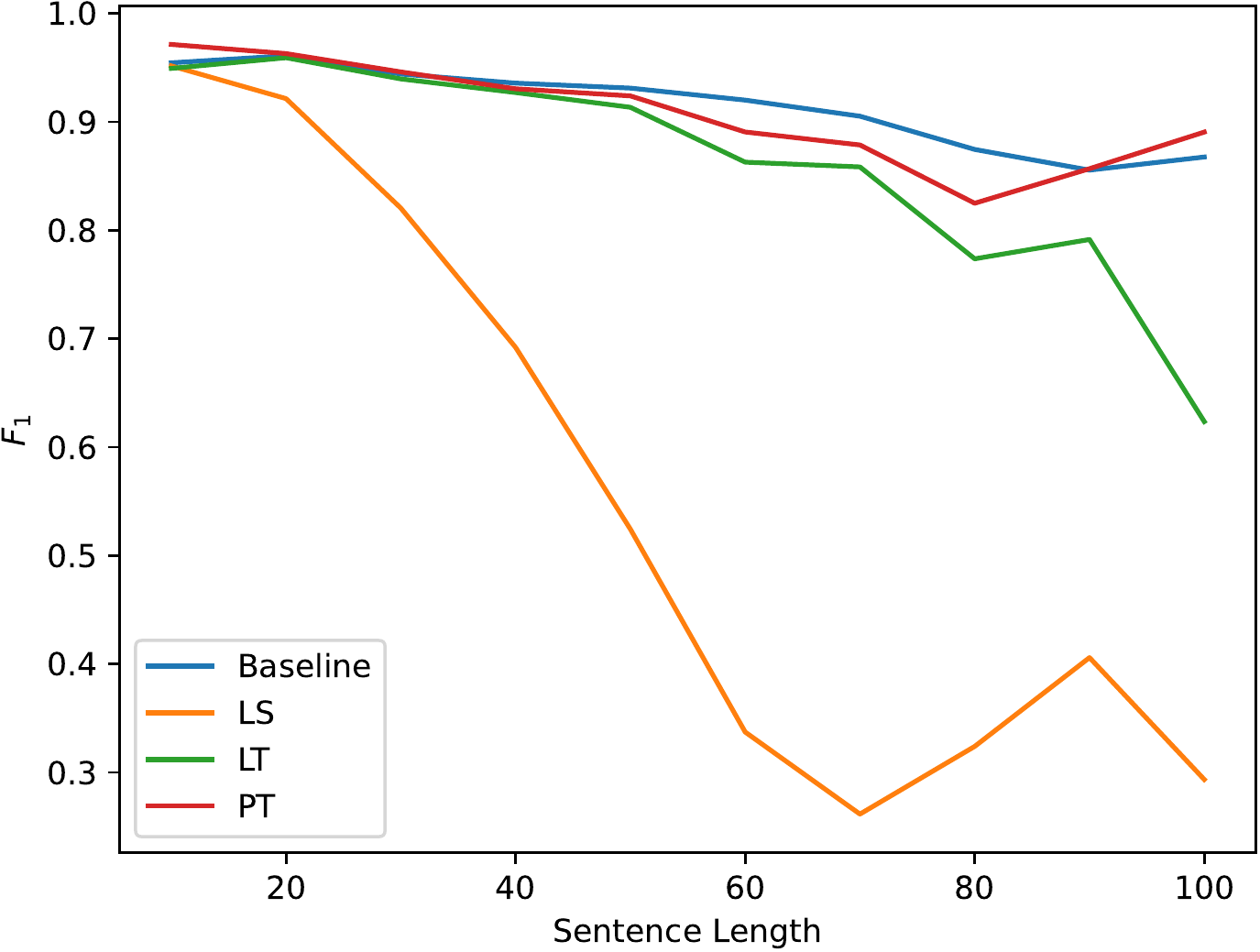}
      \caption{Length for \CON}
      \label{fig:con-length}
  \end{subfigure}
  \hfill
  \begin{subfigure}[b]{0.24\textwidth}
      \centering
      \includegraphics[width=\textwidth]{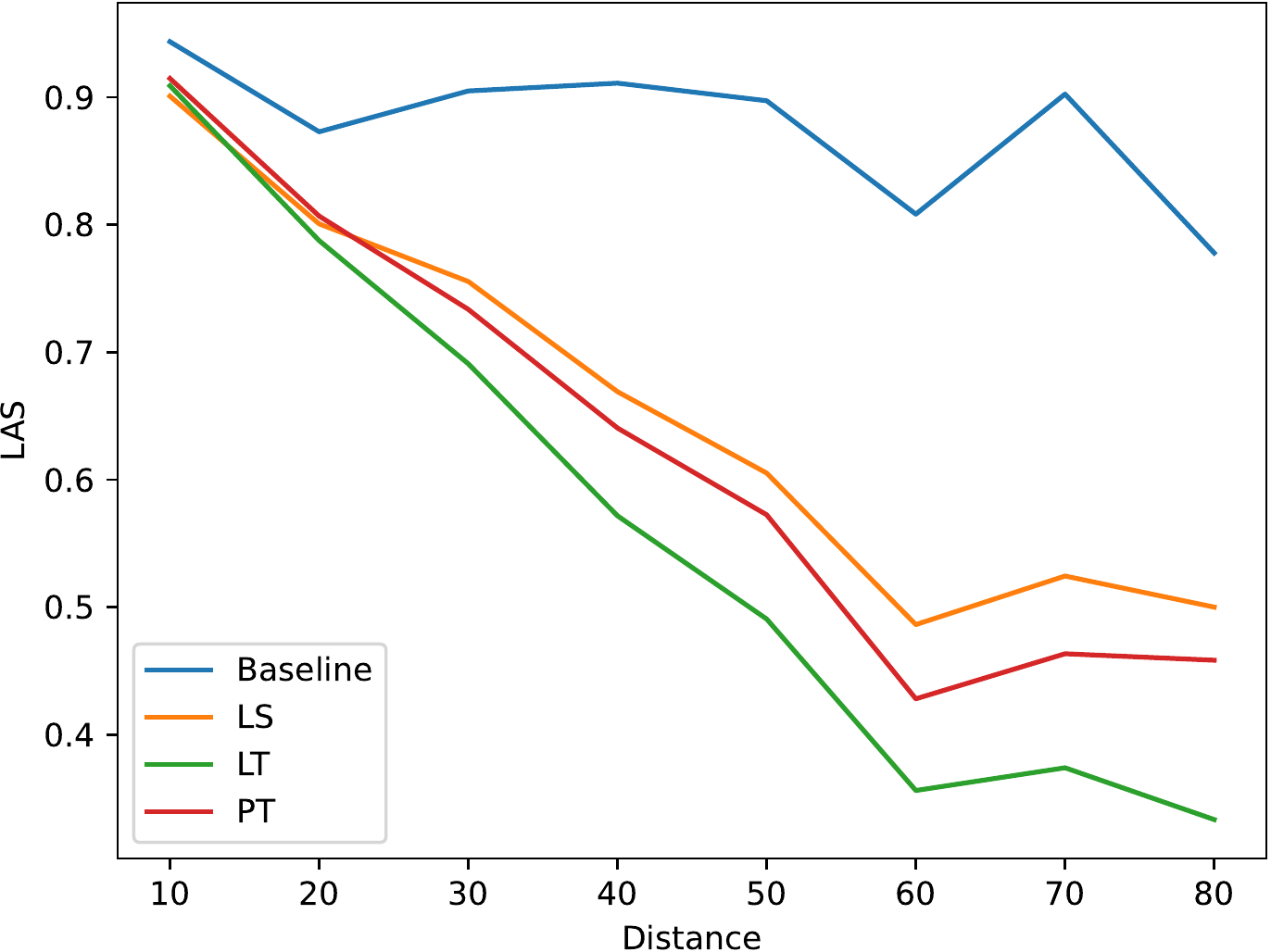}
      \caption{Distance for \DEP}
      \label{fig:dep-distance}
  \end{subfigure}
     \caption{Factors impacting each task: the rate of OOV tokens for \POS, the rate of unseen entities for \NER, the sentence length for \CON, and the head-dependent distance for \DEP.}
     \label{fig:analysis}
\end{figure*}

\subsection{Stratified Analysis}
\label{sec:further_discussion}

Section~\ref{sec:experiments} shows that our S2S approach performs comparably to most ad-hoc models. To reveal its pros and cons, we further partition the test data using task-specific factors and run tests on them. The stratified performance on OntoNotes 5 is compared to the strong BERT baseline \cite{he-choi-2021-stem}, which is representative of non-S2S models implementing many state-of-the-art decoders.

For \POS, we consider the rate of Out-Of-Vocabulary tokens (OOV, tokens unseen in the training set) in a sentence as the most significant factor. As illustrated in Figure~\ref{fig:pos-oov}, the OOV rate degrades the baseline performance rapidly, especially when over half tokens in a sentence are OOV. However, all S2S approaches show strong resistance to OOV, suggesting that our S2S models unleash greater potential through transfer learning.

For \NER, entities unseen during training often confuse a model. This negative impact can be observed on the baseline and \texttt{LT} in Figure~\ref{fig:ner-oov}. However, the other two schemas generating textual tokens, \texttt{LT} and \texttt{PT}, are less severely impacted by unseen entities. It further supports the intuition behind our approach and agrees with the finding by \citet{shin-etal-2021-constrained}: with the output sequence being closer to natural language, the S2S model has less difficulty generating it even with unseen entities.

Since the number of binary parses for a sentence of $n+1$ tokens is the $n$th \textit{Catalan Number} \cite{church-patil-1982-coping}, the length is a crucial factor for \CON. As shown in Figure~\ref{fig:con-length}, all models, especially \texttt{LS}, perform worse when the sentence gets longer. Interestingly, by simply recalling all the lexicons, \texttt{LT} easily regains the ability to parse long sentences. Using an even more natural representation, \texttt{PT} outperforms them with a performance on par with the strong baseline. It again supports our intuition that natural language is beneficial for pretrained S2S.

For \DEP, the distance between each dependent and its head is used to factorize the overall performance. As shown in \ref{fig:dep-distance}, the gap between S2S models and the baseline increases with head-dependent distance. The degeneration of relatively longer arc-standard transition sequences could be attributed to the static oracle used in finetuning.

Comparing the three schemas across all subgroups, \texttt{LT} uses the most special tokens but performs the worst, while \texttt{PT} uses zero special tokens and outperforms the rest two. It suggests that special tokens could harm the performance of the pretrained S2S model as they introduce a mismatch between pretraining and finetuning. With zero special tokens, \texttt{PT} is most similar to natural language, and it also introduces no extra parameters in finetuning, leading to better performance.

%% file: tex/conclusion.tex
\section{Conclusion}

We aim to unleash the true potential of S2S models for sequence tagging and structure parsing. 
To this end, we develop S2S methods that rival state-of-the-art approaches more complicated than ours, without substantial task-specific architecture modifications.
Our experiments with three novel prompting schemas on four core NLP tasks demonstrated the effectiveness of natural language in S2S outputs. 
Our systematic analysis revealed the pros and cons of S2S models, appealing for more exploration of structure prediction with S2S.

Our proposed S2S approach reduces the need for many heavily-engineered task-specific architectures. 
It can be readily extended to multi-task and few-shot learning. 
We have a vision of S2S playing an integral role in more language understanding and generation systems.
The limitation of our approach is its relatively slow decoding speed due to serial generation. This issue can be mitigated with non-autoregressive generation and model compression techniques in the future.